\documentclass[runningheads]{llncs}

\usepackage{graphicx}
\usepackage{comment}
\usepackage{amsmath,amssymb} 
\usepackage{color}
\usepackage{dsfont}
\usepackage{xcolor}
\usepackage{subfig}
\newcommand{\norm}[1]{\left\lVert#1\right\rVert}
\newcommand{\keypoint}[1]{\vspace{0.2cm}\noindent\textbf{#1}\quad}
\newcommand{\keypointZ}[1]{\vspace{0.0cm}\noindent\textbf{#1}\quad}

\newcommand{\cut}[1]{}

\newcommand{\bModel}{B\'ezierSketch}
\newcommand{\bEnc}{B\'ezierEncoder}

\begin{document}
\pagestyle{headings}
\mainmatter
\def\ECCVSubNumber{5496}  

\title{B\'ezierSketch: A generative model for scalable vector sketches} 

\titlerunning{ECCV-20 submission ID \ECCVSubNumber} 
\authorrunning{ECCV-20 submission ID \ECCVSubNumber} 
\author{Anonymous ECCV submission}
\institute{Paper ID \ECCVSubNumber}

\titlerunning{B\'ezierSketch: A generative model for scalable vector sketches}

\author{
Ayan Das\inst{1,2}\and
Yongxin Yang\inst{1,2}\and
Timothy Hospedales\inst{1,3}\and
Tao Xiang\inst{1,2}\and
Yi-Zhe Song\inst{1,2}
}

\authorrunning{A. Das et al.}

\institute{
SketchX, CVSSP, University of Surrey, United Kingdom \\
\email{\{a.das,yongxin.yang,t.xiang,y.song\}@surrey.ac.uk} \and
iFlyTek-Surrey Joint Research Centre on Artificial Intelligence \and
University of Edinburgh, United Kingdom \\
\email{t.hospedales@ed.ac.uk}
}
\maketitle

\begin{abstract}
The study of neural generative models of human sketches is a fascinating contemporary modeling problem due to the links between sketch image generation and the human drawing process. The landmark SketchRNN provided breakthrough by sequentially generating sketches as a sequence of waypoints. However this  leads to low-resolution image generation, and failure to model long sketches. In this paper we present \bModel{}, a novel generative model for fully \emph{vector} sketches that are automatically scalable and high-resolution. To this end, we first introduce a novel inverse graphics approach to stroke embedding that trains an encoder to embed each stroke to its best fit B\'ezier curve. This enables us to treat sketches as short sequences of paramaterized strokes \cut{(rather than pixels)}and thus train a recurrent sketch generator with greater capacity for longer sketches, while producing scalable high-resolution results. We report qualitative and quantitative results on the \emph{Quick, Draw!} benchmark. 
\end{abstract}
\vspace{-0.7cm}
\begin{figure}
    \centering
    \includegraphics[width=\linewidth]{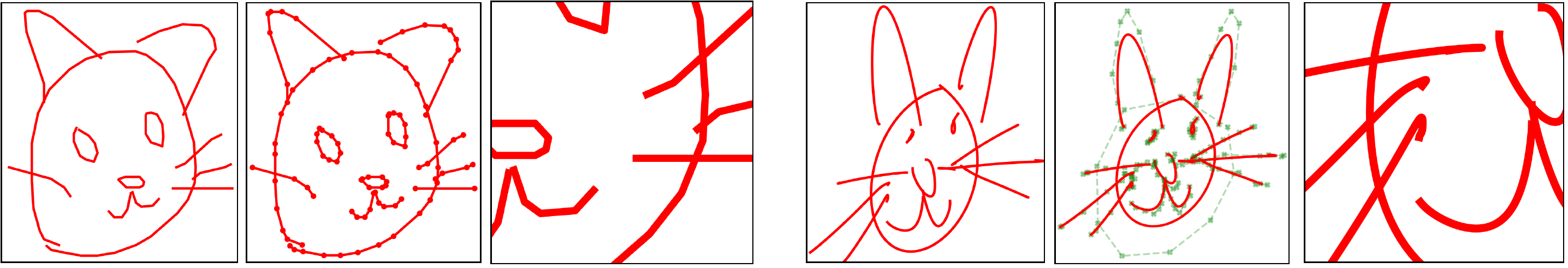}
    \vspace{-0.5cm}
    \caption{Left: SketchRNN \cite{ha2017neural} generates sketches by sampling waypoints (red dots) which lead to coarse images upon zoom. Right: Our \bModel{} samples smooth curves (green control points) thus providing scalable \emph{vector} graphic generation.}
    \label{fig:banner}
\end{figure}
\vspace{-0.7cm}

\section{Introduction}
Generative neural modeling of images \cite{GANgoodfellow14,Kingma2013AutoEncodingVB} is now an established research area in contemporary machine learning and computer vision. Rapid progress has been made in generating photos \cite{pix2pix_isola,dcgan}, with effort being focused on fidelity, diversity, and resolution of image generation, along with stability of training; as well as sequential models for text and video \cite{bowman-etal-2016-generating,srivastava2015unsupervised}.
\cut{Effort has also been directed towards sequential generative models \cite{bowman-etal-2016-generating,srivastava2015unsupervised}, primarily targeted at text and video synthesis.} Generative modeling of human \emph{sketches} in particular has recently gained interest, along with other applications of sketch analysis such as recognition \cite{sketchanet1,sketchanet2}, retrieval  \cite{sangkloy2016sketchy,sketchx_fgsbir_1,sketchx_fgsbir_2} and forensics \cite{klare2011mugshotSketchMatch} -- all facilitated by the growth of large scale sketch datasets \cite{ha2017neural,sangkloy2016sketchy}. 

Sketch generation provides an excellent opportunity to study sequential generative models, and is particularly fascinating due to the potential to establish links between learned generative models and human sketching -- a  communication modality that comes innately to children, and has existed for millennia. Recent breakthroughs in this area include SketchRNN \cite{ha2017neural}, which provided the first neural generative sequential model for sketch images, and Learn2Sketch \cite{song2018learn2Sketch} which provided the first conditional image to sequential sketch model. While conventional image generation models focus on producing ever-larger pixel arrays in high fidelity, these methods aim to model sketches using a more human-like representation consisting of a collection of strokes.

SketchRNN \cite{ha2017neural}, the landmark neural sketch generation algorithm, treats sketches as a digitized sequence of 2D points on a drawing canvas sampled along the trajectory of the ink-flow. This model of sketches has several issues, however: It is inefficient, due to the dense representation of redundant information like highly correlated temporal samples; and as  sketches are ultimately pixels on a grid, it is prone to sampling noise. Crucially it provides limited graphical scalability: SketchRNN sets out to achieve vector graphic generation (and claims to achieve this). However it does not generate truly scalable vector graphs as required by applications such as digital art. Since generated sketches are composed of dense line segments, its samples are only somewhat smoother than raster graphics (Fig.~\ref{fig:banner}).  Finally, it suffers from limited capacity. Because it models sketches as a sequence of pixels, it is limited in the length of sketch it can model before the underlying recurrent neural network begins to run out of capacity. 


In this paper we propose a fundamental paradigm change in the representation of sketches that enables the above issues to be addressed. Specifically, we aim to represent sketches in terms of parameterized smooth curves \cite{salomon2007curves}. These provide a scalable representation of a finite length curve using few \emph{Control Points}. From a large family of parametric curves, we choose B\'ezier curves due to their simple structure\cut{ and low dimensionality}. In order to train a generative model of human sketches with this representation, the key question is how to encode human sketches as parameterized curves. To this end, a key technical contribution is a vision-as-inverse-graphics \cite{kulkarni2015vig,romaszko2017vig,ganin2018imagePrograms} approach, that learns to embed human sketch strokes as interpretable parameterized B\'ezier curves. We train \bEnc{} in an inverse-graphics manner by learning to reconstruct strokes through a white-box graphics (B\'ezier) decoder. Given this new low-dimensional stroke representation, we then train \bModel{} to generate sketches. Our stroke-level generative model requires many fewer iterations than the segment-level SketchRNN, and thus provides better generation of longer sketches, while providing high-resolution scalable  vector-graphic sketch generation (Fig.~\ref{fig:banner}).

In summary, the contributions of our work are: (1) \bEnc{}, a novel inverse-graphics approach for mapping strokes to parameterized B\'eziers, (2) \bModel{}, a sequential generative model for sketches that produces high-resolution  and low-noise vector graphic samples with improved scalability to longer sketches compared to the previous state of the art SketchRNN.


\section{Related Work}

\keypointZ{Parameterized Curves}
B\'ezier curves are a powerful tool in the field of computer graphics and are extensively used in interactive curve and surface design \cite{salomon2007curves}, as are a more general family of curves known as \emph{Splines}  \cite{deboor}. Optimization algorithms to fit B\'ezier curves and Splines from data have been studied. Few specially crafted algorithms do exist specifically for cubic B\'ezier curves \cite{cubicbezierfit,advancecubicbezierfit}. 
However the challenge for most curve and spline-fitting methods is the existence of latent variables $t$ that correspond training points and the location of their projection onto the curve. This leads to two-stage alternating algorithms for separately optimizing the curve parameters (control points) and latent parameter $t$ \cite{spliefit_persample1,spliefit_persample2}. Importantly, such methods \cite{spliefit_persample1,spliefit_persample2} including few promising ones \cite{bspline_lbfgs} require expensive \emph{per-sample} alternating optimization, or iterative inference in expensive generative models \cite{digit_splines_hinton,lake_bpl} which make them unsuitable for large scale or online applications. In contrast, we uniquely take the approach of learning a neural network that maps strokes to B\'ezier curves in a single shot. This neural encoder is a model that needs to be trained, but unlike per-sample optimization approaches, it is inductive. So once trained it can provide one-shot estimation of curve parameters and point association from an input stroke.

\keypointZ{Generative Models}
Generative models have been studied extensively in the machine learning literature, often in terms of density estimation with directed \cite{rabiner_hmm,bishop1994mixture} or undirected \cite{Hinton504} graphical models. Research in this field accelerated after the emergence of Generative adversarial networks (GAN) \cite{GANgoodfellow14}, Variational Autoencoder (VAE) \cite{Kingma2013AutoEncodingVB} and their derivatives. Handling sequences are of particular importance and hence specialized algorithms \cite{bowman-etal-2016-generating,srivastava2015unsupervised} were developed. Although RNNs have been successfully used for generating handwriting \cite{graves13} without variational training, these methods lacked flexibility in terms of generation quality. The emergence of VAE and variational training methods allows the fusion of RNNs with variational objective led to the first successful generative sequence model \cite{bowman-etal-2016-generating} in the domain of Natural Language Processing (NLP). It was quickly adapted by SketchRNN \cite{ha2017neural} in order to extend \cite{graves13} to free-hand sketches.

\keypointZ{Inverse Graphics} ``Inverse Graphics" is line of work that aims to estimate 3D scene parameters from raster images without supervision. Instead it predicts the input parameters of a computer graphics pipeline that can reconstruct the image. Several attempts were made \cite{romaszko2017vig,kulkarni2015vig} to estimate explicit model parameters of 3D objects from raw images. A specialized case of the generic Inverse Graphics idea is to estimate parameters of 2D objects such as curves. As a recent example, an RNN based agent named SPIRAL \cite{ganin2018imagePrograms} learned to draw characters in terms of pen an brush curves. SPIRAL, however, is extremely costly due to its reliance on Policy Gradient \cite{policygrad} reinforcement learning training and black-box renderer.
 
\keypointZ{Learning for Curves}
Few works have studied learning for curve generation. The recent SVG Font Generator \cite{fontgen_iccv} trains an excellent font embedding with a recurrent vector font image generator. However it is trained with supervision rather than inverse graphics, and limited to the more structured domain of font images. Other attempts \cite{laube2018bSplineNeural} also use supervised learning on synthetic data, rather than unsupervised learning on real human sketches as we consider here.

\section{Methodology}
\keypoint{Background: Conventional Sketch representation and Generation}
A common format \cite{ha2017neural} for a digitally acquired sketch $\mathcal{S}$ is as a sequence of 2-tuples, each containing a $2D$ coordinate on the canvas sampled from a continuous drawing flow and a pen-state bit denoting whether the pen touches the canvas or not. 

\begin{equation} \label{eq:OriginalDS}
\mathcal{S} = \bigl[ \left( \mathbf{X}_i, q_i \right) \bigr]_{i=1}^L
\end{equation}

\noindent where $\mathbf{X}_i \triangleq \bigl[ x\ y \bigr]_i^T \in \mathbb{R}^2 $, $q_i \in \{ \textsc{PenUp}, \textsc{PenDown} \}$ and $L$ is the cardinality of $\mathcal{S}$ representing the length of the sketch. The state-of-the-art sketch generator SketchRNN \cite{ha2017neural} learns a parametric Recurrent Neural Network (RNN) to model the joint distribution of  coordinates and pen state as a product of conditionals, i.e. 
$p_{sketchrnn}(\mathcal{S}; \theta) = \prod_{i=1}^L p\bigl(\mathbf{X}_i, q_i \left| \mathbf{X}_{<i}, q_{<i}; \theta \right. \bigr)$, where $\theta$ is the set of parameters of the model and $\mathbf{X}_{<i}$ and $q_{<i}$ denote the list of locations and pen-state bits respectively before $\mathbf{X}_i$ and $q_i$. 

\keypoint{Towards a Stroke-Level Representation} We are interested in moving from such a segment-level representation toward stroke-level. To this end we modify the structure of our input data  to $\bar{\mathcal{S}} \triangleq \bigl[ \mathbf{T}_j \bigr]_{j=1}^N \text{, with } \mathbf{T}_j \triangleq \bigl[ \mathbf{X}_i^{(j)}\bigr]_{i=1}^{N_j}$ where $\mathbf{T}_j$ is the $j^{th}$ stroke of length $N_j \triangleq \vert \mathbf{T}_j \vert$ segregated from the sketch by following the pen-state bit, and consequently $\sum_{j=1}^N N_j = L$. 

\keypoint{Towards a Stroke-Level Generative Model} Existing generative sketch models \cite{ha2017neural,song2018learn2Sketch} generate a segment at each iteration. Given a stroke-segmented training set  $\bar{\mathcal{S}}$, we would like to train a generative model analogous to SketchRNN. That is, to model the distribution over possible sketches with a parametric model $p_{model}(\bar{\mathcal{S}}; \theta)$ and that approximates the original data distribution $p_{data}(\bar{\mathcal{S}})$. Different sketches having different lengths $N$ makes this problem suitable for Recurrent Neural Networks (RNN). One could model the probability of a sketch as a product of the probabilities of individual strokes $\mathbf{T}_j$ conditioned on all its previously seen strokes $\mathbf{T}_{< j}$ and parameterized by set of parameters $\theta$ as $p_{model}(\bar{\mathcal{S}}; \theta) = \prod_j p(\mathbf{T}_j \vert \mathbf{T}_{< j}; \theta)$. However, a  problem with such an approach is that the individual strokes $\mathbf{T}_j$ are of varying length which would require a hierarchical model where $p(\mathbf{T}_j\vert \cdot)$ is again modeled as a sequence. So we instead propose to learn fixed length  embedding $\mathbf{e}_j \triangleq \mathbf{e}(\mathbf{T}_j) \in \mathbb{R}^d$ for any stroke $\mathbf{T}_j$ {and corresponding non-parametric decoder $\mathbf{d}(\cdot)$ such that $\mathbf{T}_j \approx \mathbf{d}(\mathbf{e}_j)$. We then model the encoded sketch $\mathbf{e}(\bar{\mathcal{S}}) \triangleq \bigl\{ \mathbf{e}_j \bigr\}_{j=1}^N$ as}

\begin{equation} \label{eq:StrokeRNN}
p_{model}(\mathbf{e}(\bar{\mathcal{S}}); \theta)  = \prod_{j=1}^N p(\mathbf{e}_j \vert \mathbf{e}_{< j}; \theta)
\end{equation}

\noindent where the individual conditionals are typically one or more mixtures of Gaussians (GMMs) and where the raw sketch can be rendered at any point by the decoder. In order to sample a new sketch from the model, we sample each $j^{th}$ stroke from $p(\mathbf{e}_j \vert \mathbf{e}_{< j}; \theta)$ \cut{by conditioning them on all previously sampled strokes} and render it as $\mathbf{d}(\mathbf{e}_j)$. 

A natural choice for the embedding $\mathbf{e}(\cdot)$ could be an encoder RNN trained as part of a Sequence-to-Sequence autoencoder \cite{srivastava2015unsupervised}. However, We take a different approach and propose a novel inverse-graphics based encoder-decoder framework $\mathbf{T}\approx \mathbf{d}(\mathbf{e}(\mathbf{T}))$ where our neural  encoder $\mathbf{e}(\cdot)$ produces an \emph{interpretable} representation because it must decode through a white-box B\'ezier renderer $\mathbf{d}(\cdot)$. 



\subsection{Stroke embedding: \bEnc{}} \label{sec:StrokeEmbedding}

To train our parametric stroke embedding with an inverse graphics strategy, we must first define a differentiable `graphics decoder' which will be later used to train our neural encoder to map human strokes to B\'ezier curves.
\cut{We embed each stroke $\mathbf{T}_j$ with varying length $N_j$ as B\'ezier curves \cite{salomon2007curves} with pre-defined degree $n$.}

\keypoint{Inverse Graphics Decoder}
B\'ezier curves, used heavily in computer graphics, are smooth curves representable in a closed functional form parameterized by a sequence of $n+1$ anchor coordinates $\mathbf{P} \triangleq \bigl[ P_x\ P_y \bigr]^T \in \mathbb{R}^2$ termed \emph{control points}. A degree $n$ B\'ezier curve with control points $\bigl[ \mathbf{P}_0, \mathbf{P}_1, \cdots \mathbf{P}_n \bigr]$ is represented as

\begin{equation} \label{eq:BezierEq}
\mathbf{C}(t; \left\{ \mathbf{P}_i \right\}) = \sum_{i=0}^n \mathcal{B}_{i,n}(t) \cdot \mathbf{P}_i
\end{equation}

\begin{figure}[t]
    \centering
    \subfloat[]{{ \includegraphics[width=0.23\linewidth]{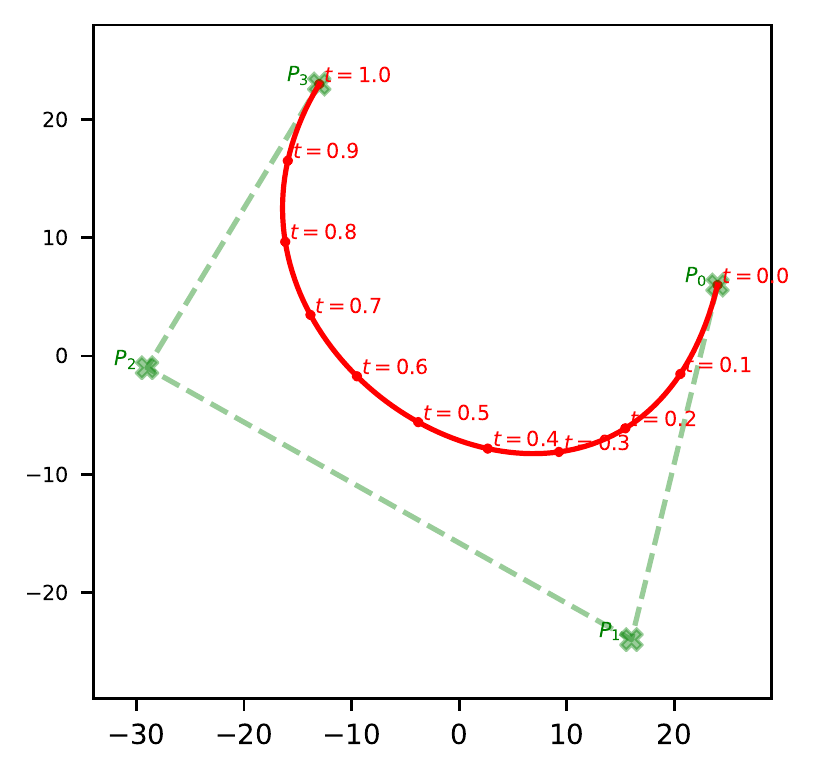} }}%
    \qquad
    \subfloat[]{{ \includegraphics[width=0.68\linewidth]{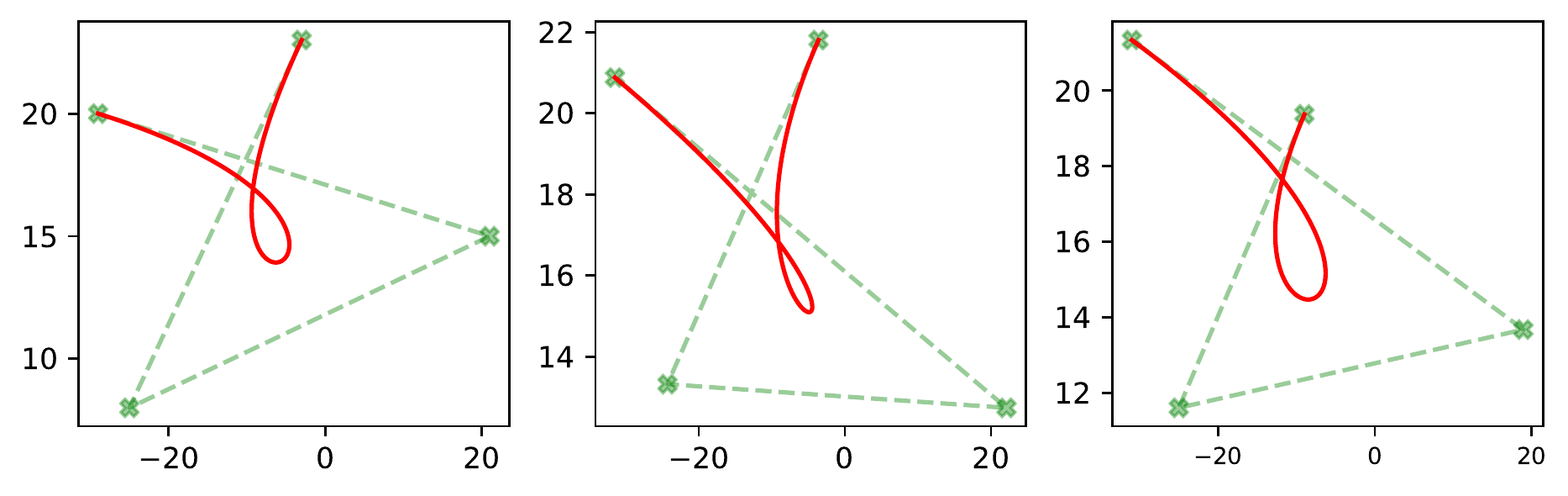} }}
    \caption{(a) An example of B\'ezier curve of degree $n=3$ with $n+1$ control points. (b) B\'ezier curves with Gaussian noise ($\mathbf{\mu}=\mathbf{0}, \Sigma=5I_2$) added to control points produce similar curves in image space.}
    \label{fig:BezierCurve}
\end{figure}

\noindent where $t \in [0, 1]$ is the \emph{parameter} of the curve, $\mathcal{B}_{i,n}(t) \triangleq \binom{n}{i} t^i (1-t)^{n-i}$ is the Bernstein Basis Polynomial in $t$ and $\mathbf{C}(t) \triangleq \bigl[ \mathrm{C}_x(t)\ \mathrm{C}_y(t) \bigr]^T \in \mathbb{R}^2$ denotes a point on the curve at $t=t$. As $t$ assumes values $0 \rightarrow 1$, the curve starts from $\mathbf{P}_0$ and ends at $\mathbf{P}_n$ and the control points $\bigl[ \mathbf{P}_1, \cdots, \mathbf{P}_{n-1} \bigr]$ control the trajectory of the curve, as illustrated in Fig.~\ref{fig:BezierCurve}(a). We further use $\mathcal{P}^n \triangleq \bigl[ {P_x}_0, {P_y}_0, \cdots, {P_x}_n, {P_y}_n \bigr] \in \mathbb{R}^{2(n+1)}$ to denote elements (curves) in the continuous space of $n+1$ control points.  The decoder function $\mathbf{d} : \mathcal{P} \rightarrow \mathbf{T}$ can be trivially realized by Eq.~\ref{eq:BezierEq} with the set of $t$-values chosen as per \cut{scalability}resolution requirement. 

We now  denote ($\mathbf{T}, \mathcal{P}$) as an arbitrary stroke and its B\'ezier representation, where we have dropped the subscript $j$ and superscript $n$ for notational brevity. Using  $\mathcal{P}$ as an embedding space for $\mathbf{T}$ leads to an extremely useful and key property: Given a choice of $n$, two similar points in $\mathcal{P}$ space correspond to similar strokes in $\mathbf{T}$ space. As a consequence, we can sample from the conditionals in Eq.~\ref{eq:StrokeRNN} to generate variations of a stroke.

\begin{property} \label{prop:SampleBezier}
Given a $(\mathbf{T}, \mathcal{P})$ pair where $\mathbf{T}=\mathbf{d}(\mathcal{P})$ and sample $\widehat{\mathcal{P}} \sim \mathcal{N}(\mathcal{P}, \sigma)$, then the decoded $\widehat{\mathbf{T}} = \mathbf{d}(\widehat{\mathcal{P}})$ is distributed as $\mathcal{N}(\mathbf{T}, \sigma')$.
\end{property}

\begin{proof}
Refer to Appendix A in the supplementary document for the proof. Illustrative examples are given in Fig.~\ref{fig:BezierCurve}(b).
\end{proof}

\cut{As is evident from Eq.~\ref{eq:BezierEq}, the solution of $\mathcal{P}^n$ is unique given a set of curve points $\left[ \mathbf{X}_i^{(j)}\right]_{i=1}^{N_j} \equiv \mathbf{T}_j$ and corresponding t-values $\left[ t_i \right]_{i=1}^{N_j}$, we choose to have our embedding function as $\mathbf{e}: \mathbf{T} \rightarrow \mathcal{P}$.}

\keypoint{A stroke to B\'ezier encoder}
We wish to learn an embedding function $\mathbf{e}(\cdot)$ that will map a given stroke $\mathbf{T}$ to its {best fit} B\'ezier representation  $\mathcal{P}$. Due to the variable length of strokes $\mathbf{T}$, we model \bEnc{} with a bi-directional RNN, with  forward and backward states $\overrightarrow{\mathbf{s}_i}, \overleftarrow{\mathbf{s}_i} \in \mathbb{R}^h$ at time-step $i$ as

\begin{equation} \label{eq:EncoderRNNTheta}
\left[ \overrightarrow{\mathbf{s}_i}, \overleftarrow{\mathbf{s}_i} \right] = \mathrm{BiRNN}(\mathbf{X}_{i-1}, \mathbf{s}_{i-1}; \theta)
\end{equation}

However, unlike regular encoder RNNs, we further transform the last hidden state to get a B\'ezier curve representation

\begin{equation} \label{eq:BazierAEInference}
{\mathcal{P}} = \mathbf{W}_{\mathcal{P}} \left[ \overrightarrow{\mathbf{s}}_{end}; \overleftarrow{\mathbf{s}}_{end} \right]
\end{equation}

\noindent where the `$end$' subscript denotes the state of the RNN at last time-step, $\left[\ ;\ \right]$ denotes the concatenation operator and $\mathbf{W}_{\mathcal{P}} \in \mathbb{R}^{2(n+1) \times 2h}$. 

\begin{figure}[t]
    \centering
    \includegraphics[scale=0.8]{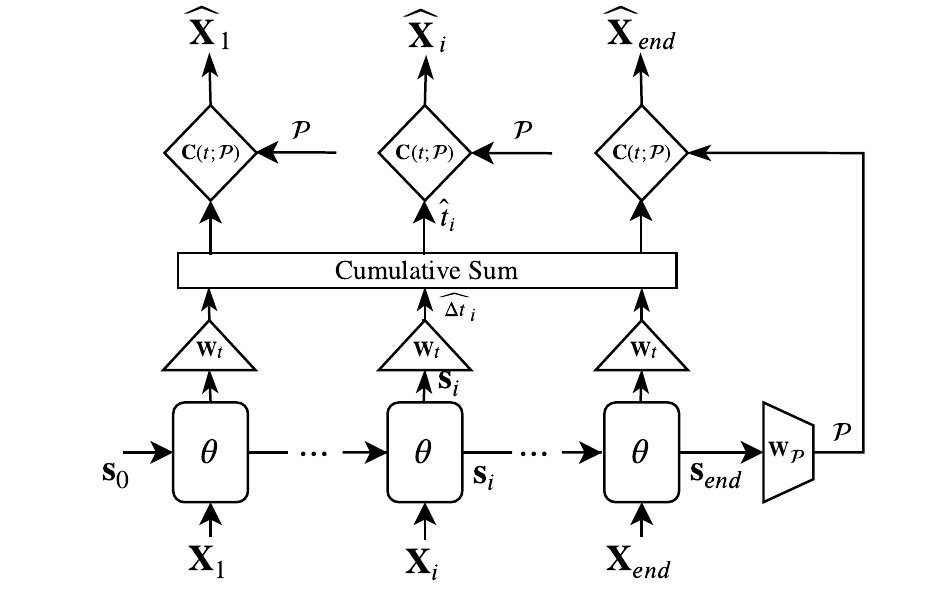}
    \caption{Inverse graphics training of our \bEnc{} architecture for model-based single-pass stroke $\left[ \mathbf{X}_i \right]$ to B\'ezier $\mathcal{P}$ mapping.}
    \label{fig:BezierAE}
\end{figure}

The formulation so far enables extracting a curve $\mathcal{P}$ from data $\mathbf{T}$. However, while $\mathcal{P}$ is now a sufficient representation to decode the B\'ezier by means of Eq.~\ref{eq:BezierEq}, we do not have sufficient information to compute a reconstruction loss like  $\|\mathbf{T}-\mathbf{d}(\mathbf{e}(\mathbf{T}))\|$ because we lack the association between input coordinates $\mathbf{X}_i$  and interpolation parameters $t_i$. This is where many classic B\'ezier fitting techniques \cite{spliefit_persample1,bspline_lbfgs} resort to slow alternating optimization techniques. 

We take a different approach and ask our encoder to also predict the corresponding interpolation parameter $t_i$ for each input point $\mathbf{X}_i$. \cut{Specifically, we assume $t_i$ values are predictable given $\mathbf{X}_i$ and $\mathcal{P}$ as $\widehat{t}_i = \textsc{Sigmoid}(F(\mathbf{X}_i, \mathcal{P}))$.} In order to make valid predictions for $t$ we note the properties it requires due to its role in B\'ezier curves generation: \textbf{1.} $0 \leqslant \widehat{t}_i \leqslant 1$ (by definition of B\'ezier curve). \textbf{2.} $\widehat{t}_i \leqslant \widehat{t}_{i+1}$ (due to sequential nature of $\mathbf{X}_i$). Apart from these, we impose another property without any lose of generality: \textbf{3.} $t_1 = 0$ and $t_{end} = 1$ (this will make $\mathbf{X_1}$ and $\mathbf{X}_{end}$ coincide with $\mathbf{P}_0$ and $\mathbf{P}_n$ respectively). Please refer to the experiment section for an implementation trick to do so. 

To enable our encoder to meet these requirements above, we do not compute $t_i$s directly, but instead compute increments $\Delta t_i$ \cut{as $\Delta t_i$}$\triangleq t_i - t_{i-1}$ (with $t_0 \triangleq 0$) from $\left[ \overrightarrow{\mathbf{s}_i}; \overleftarrow{\mathbf{s}_i} \right]$ at every step $i$.
The $t_i$-values can then be easily computed as a cumulative sum of all $\Delta t_i$ up to $i$. Thus, the second path of our encoder predicts

\begin{equation} \label{eq:tEstimate}
\widehat{t}_i = \sum_{i'=1}^i \widehat{\Delta t}_{i'}\text{, with } \widehat{\Delta t}_i = \textsc{Softmax}_i(\mathbf{W}_t \cdot \left[ \overrightarrow{\mathbf{s}_i}; \overleftarrow{\mathbf{s}_i} \right]).
\end{equation}

\noindent The usage of \textsc{Softmax()} enforces all three requirements stated above.

To summarize: Our full architecture, as shown in Figure~\ref{fig:BezierAE} thus has two pathways: A B\'ezier embedding pathway that predicts the curve $\mathcal{P}$ for the entire stroke input $\mathbf{T}$ and an interpolation parameter pathway that further predicts the estimated curve parameter $\widehat{t}_i$ for each input point $\mathbf{X}_i$ in $\mathbf{T}$. Given the ($\mathbf{X}_i, \widehat{t}_i$) pairs and $\mathcal{P}$ predicted by our encoder, we can now train our model with the following reconstruction loss:

\begin{equation} \label{eq:BezierLoss}
\mathcal{L}(\theta, \mathbf{W}_{\mathcal{P}}, \mathbf{W}_t) \triangleq \sum_i \norm{ \mathcal{C}(\widehat{t}_i, \mathcal{P}) - \mathbf{X}_i }^2
\end{equation}

\noindent which is optimized w.r.t. encoder parameters $\left\{ \theta, \mathbf{W}_{\mathcal{P}}, \mathbf{W}_t \right\}$ by SGD. Once trained, we can  compute the best-fit B\'ezier for any stroke using Eq.~\ref{eq:BazierAEInference}, which provides a  feed-forward single pass solution to a typically alternating optimization.

\keypoint{A Multi-Degree Representation Extension}
To add more flexibility, we can extend this basic building block to learn a multi-degree representation of a given stroke $\mathbf{T}$. In order to do so, we encode the stroke using the the same RNN in Eq.~\ref{eq:EncoderRNNTheta} parameterized by $\theta$ but use a set of different $\mathbf{W}_{\mathcal{P}^n}$ and $\mathbf{W}_t^n$ for a predefined range of degree $n\in \left[ n_{min}, \cdots, n_{max} \right]$ to predict B\'ezier representations of different degrees along with their corresponding $t_i^n$-values. 

\begin{equation}
\begin{split}
    \widehat{t}^n_i = \sum_{i'=1}^i \widehat{\Delta t}^n_{i'}\text{, with } \widehat{\Delta t}^n_i &= \textsc{Softmax}_i(\mathbf{W}^n_t \cdot \left[ \overrightarrow{\mathbf{s}_i}; \overleftarrow{\mathbf{s}_i} \right]) \text{ and}\\
    \mathcal{P}^n &= \mathbf{W}_{\mathcal{P}^n} \left[ \overrightarrow{\mathbf{s}}_{end}; \overleftarrow{\mathbf{s}}_{end} \right]
\end{split}
\end{equation}

The total loss is now the sum of losses at every order $n$:


\begin{equation} \label{MultiBezierLoss}
\mathcal{L}_{total} \triangleq \sum_{n=n_{min}}^{n_{max}} \mathcal{L}_n\text{, with }\mathcal{L}_n(\theta, \mathbf{W}_{\mathcal{P}^n}, \mathbf{W}^n_t) \triangleq \sum_i \norm{ \mathcal{C}(\widehat{t}^n_i, \mathcal{P}^n) - \mathbf{X}_i }^2
\end{equation}

Inference in this model can now predict a \emph{set} of B\'ezier representations for different degrees, where higher order curves fit the data better at the cost of more control points. The preferred order can then be chosen manually according to user requirement, or automatically by heuristic. An effective heuristics is to evaluate the loss $\mathcal{L}_n$ for all $n$ and choose the smallest $n$ for which $\mathcal{L}^n \leq L_{tolerance}$.

\keypoint{Smoothness Regularizer}
Our training objectives Eq.~\ref{eq:BezierLoss} or Eq.~\ref{MultiBezierLoss} may lead to overfitting in the domain of B\'ezier curves during encoder learning. To avoid this we add a smoothness regularizer (with regularization strength $\beta$) that prefers sequential control points to be nearby. Specifically, we add $\beta\cdot \mathcal{R}_n$ with $\mathcal{L}_n$ for each $n$, where $\displaystyle{ \mathcal{R}_n(\mathcal{P}^n) \triangleq \sum_{i=1}^n \norm{\mathbf{P}_{i+1} - \mathbf{P}_i}_2^2 }$.


\subsection{Sketch generation: \bModel{}} \label{sec:SketchGen}

We next leverage our choice of B\'ezier representation space, and encoding model $\mathcal{P}=\mathbf{e}(\cdot)$ to define two alternative vector graphic generative models for sketches.

\keypoint{Control Point mode}
Given a sketch as a sequence of stroke embeddings $\left\{ \mathcal{P}_j \right\}_{i=1}^N$ obtained from the raw input strokes as $\mathcal{P}=\textbf{e}(\mathbf{T})$, we can modify the original data structure in Eq.~\ref{eq:OriginalDS} and substitute the set of absolute co-ordinates of every stroke by the set of control points of its B\'ezier representation. The modified sketch $\mathcal{S}_{cp}$ would be
\begin{equation} \label{eq:ControlPointModeStructure}
\mathcal{S}_{cp} = \left[ \left( \mathbf{P}^{(j)}_0, q^{(j)}_0 \right), \cdots, \left( \mathbf{P}^{(j)}_i, q^{(j)}_i \right), \cdots, \left( \mathbf{P}^{(j)}_{n_j}, q^{(j)}_{n_j} \right) \right]_{j=1}^N
\end{equation}

When encoded this way by our B\'ezier encoder, each sketch is represented by a relatively shorter (mostly) list of parametric control points rather than the original long list of coordinates. In this format, different strokes can have different degrees, as indicated by the use of $n_j$ above. \cut{The total number of  elements in the above structure is $\vert \mathcal{S}\vert = \sum_{j=1}^N (n_j + 1)$. We keep the definition of $q_i^{(j)}$ exactly as in Eq.~\ref{eq:OriginalDS}.}

Given this sequential representation of a sketch dataset, we can now train a generative sketch model. Since $\mathcal{S}_{cp}$ is structurally same as original $\mathcal{S}$ apart from its length and the interpretation of its co-ordinates, we can re-use exactly the same architecture and training procedure as SketchRNN \cite{ha2017neural}. We use a variational sequence-to-sequence autoencoder \cite{srivastava2015unsupervised} with a latent vector encoding the whole sketch. Thus one sketch is encoded first to a list of B\'ezier curves, and then to a latent vector in SketchRNN architecture; and decoded first to a list of curve parameters, and then rendered by the B\'ezier renderer.
Please refer to Appendix B for a brief review of the SketchRNN architecture in the context of our problem.

\keypoint{Stroke mode}
Given a sketch $\mathcal{S}$ as set of strokes $\left\{ \mathbf{T}_j \right\}_{j=1}^N$, we transform it as $\mathcal{S}_{st} = \left\{ \mathcal{P}_j \right\}_{j=1}^N$ where $\mathcal{P}_j = \mathbf{e}(\mathbf{T}_j)$. We model the whole sketch using a sequence-to-sequence autoencoder, where each time-step processes one stroke represented as a fixed order B\'ezier curve. We use a bi-directional RNN to encode the whole sketch stroke-by-stroke. The hidden states (forward and backward) of the encoder $\overrightarrow{\mathbf{h}}_j, \overleftarrow{\mathbf{h}}_j$ at time-step $j$ is given as

$$
\left[ \overrightarrow{\mathbf{h}}_j, \overleftarrow{\mathbf{h}}_j \right] = \mathrm{BiRNN}(\mathcal{P}_{j-1}, \mathbf{h}_{i-1}; \Theta)
$$

A latent vector $\mathbf{z} \in \mathbb{R}^{N_z}$ encoding the whole sketch is sampled using the parameters of a Gaussian distribution computed from the last hidden states

$$
\mathbf{z} \sim \mathcal{N}(\mu_{\mathbf{z}}, \text{diag}(\sigma_{\mathbf{z}}))\text{, with }
\left[ \mu_{\mathbf{z}}, \sigma_{\mathbf{z}} \right] = f\left(\left[ \overrightarrow{\mathbf{h}}_N; \overleftarrow{\mathbf{h}}_N \right]; \Theta\right)
$$

An unidirectional decoder RNN is initialized using $\mathbf{z}$ and models the probability of $j^{th}$ stroke embedding conditioned on the hidden state $\mathbf{g}_j \in \mathbb{R}^{H^d}$

\begin{equation} \label{eq:SketchDecoder}
\begin{split}
p(\mathcal{P}_j \vert \mathbf{g}_j; \Theta) &= \mathrm{GMM}\left(\mathcal{P}_j; \left\{ \mu^m_j( \mathbf{g}_j), \Sigma^m_j(\mathbf{g}_j), \pi^m_j(\mathbf{g}_j) \right\}_{m=1}^M\right) \\
\mathbf{g}_j &= \mathrm{DecoderRNN}(\left[ \mathcal{P}_{j-1}; \mathbf{z} \right], \mathbf{g}_{j-1}; \Theta)
\end{split}
\end{equation}

\noindent where $\left\{ \mu^m_j, \Sigma^m_j, \pi^m_j \right\}$ are the parameters of the $M$-component GMM for the $j^{th}$ stroke. For computational efficiency, we consider diagonal $\Sigma^m_j$ and by definition $\sum_m \pi^m_j = 1$. Given a trained model, we can sample from this distribution to generate similar $\mathcal{P}_j$ which will resemble its original domain data $\mathbf{T}_j$ as guaranteed by property~\ref{prop:SampleBezier}. Along with $\mathcal{P}_j$ at every step $j$, we also predict a stop bit $\widehat{b}_j \in \left[ 0, 1 \right]$ denoting end of sketch which is compared against the ground-truth stop bit $b_j \triangleq \mathds{1}_{j=N}$. The sketch generator is trained with the following objective function

\begin{equation}
\begin{split}
\mathcal{L}(\left\{ \mathcal{P}_j \right\}_{i=1}^N; \Theta) &= \left[- \frac{1}{N_{max}} \sum_{j=1}^{N} \log \mathrm{GMM}\left(\mathcal{P}_j \vert \bigl\{ \mu^m_j, \Sigma^m_j, \pi^m_j \bigr\}_{m=1}^M; \Theta\right) \right. \\
- \frac{1}{N_{max}} &\left.\sum_{j=1}^{N} b_j \log \widehat{b}_j \right] - \frac{1}{2 N_z} \sum_{i=1}^{N_z} \left( 1 + \sigma^i_{\mathbf{z}} - \mu^i_{\mathbf{z}} - exp(\sigma^i_{\mathbf{z}}) \right)
\end{split}
\end{equation}

{The first two terms of $\mathcal{L}$ are the log-likelihood of a sequence $\left\{ \mathcal{P}_j \right\}_{i=1}^N$ under the model and the loss due to the stop bit respectively.} The third term denotes the KL-divergence loss for imposing a Gaussian prior on the latent code $\mathbf{z}$. The diagonal entries of $\Sigma^m_j$ have been raised by $exp(\cdot)$ to make them non-negative and $\textsc{Softmax}(\cdot)$ has been used to ensure $\sum_m \pi^m_j = 1$.

\section{Experiments \& Results}

\keypoint{Dataset}
\emph{Quick, Draw!} is a large sketch dataset  \cite{ha2017neural} collected as a part of an online game to draw a given category within a time-limit, in which thousands of people around the world participated.  Due to the problem definition and structure of data used by our framework (see Eq.\ref{eq:OriginalDS}), \emph{Quick, Draw!} is the most suitable dataset to validate it. 
Different versions of the dataset use different sampling rates at which the sketches are stored as point sequences. SketchRNN is known to work well only on data with lower sampling rate (i.e., $\mathbb{E}_{\mathbf{T}} \bigl[ \vert \mathbf{T} \vert \bigr]$ is lower) than the raw data ($\mathbb{E}_{\mathbf{T}} \bigl[ \vert \mathbf{T} \vert \bigr]$ is higher) recorded. Due to fixed length of B\'ezier representations, our framework can adapt to data with both high and low sampling rates without any modification. Although our method is generalizable across all categories, we experimented with few categories to validate our claims.

Our framework has two main components: \textbf{1.} Embedding each stroke into its B\'ezier representation. \textbf{2.} Training a generative model with the encoded sketches either in \emph{control point mode} or \emph{stroke mode}. As our \bEnc{} is a key contribution, we validate this in isolation, before comparing  our whole \bModel{} framework to SketchRNN \cite{ha2017neural}.

\subsection{Stroke Embedding Experiments}


\begin{figure}[t]
    \centering
    \includegraphics[width=0.57\linewidth]{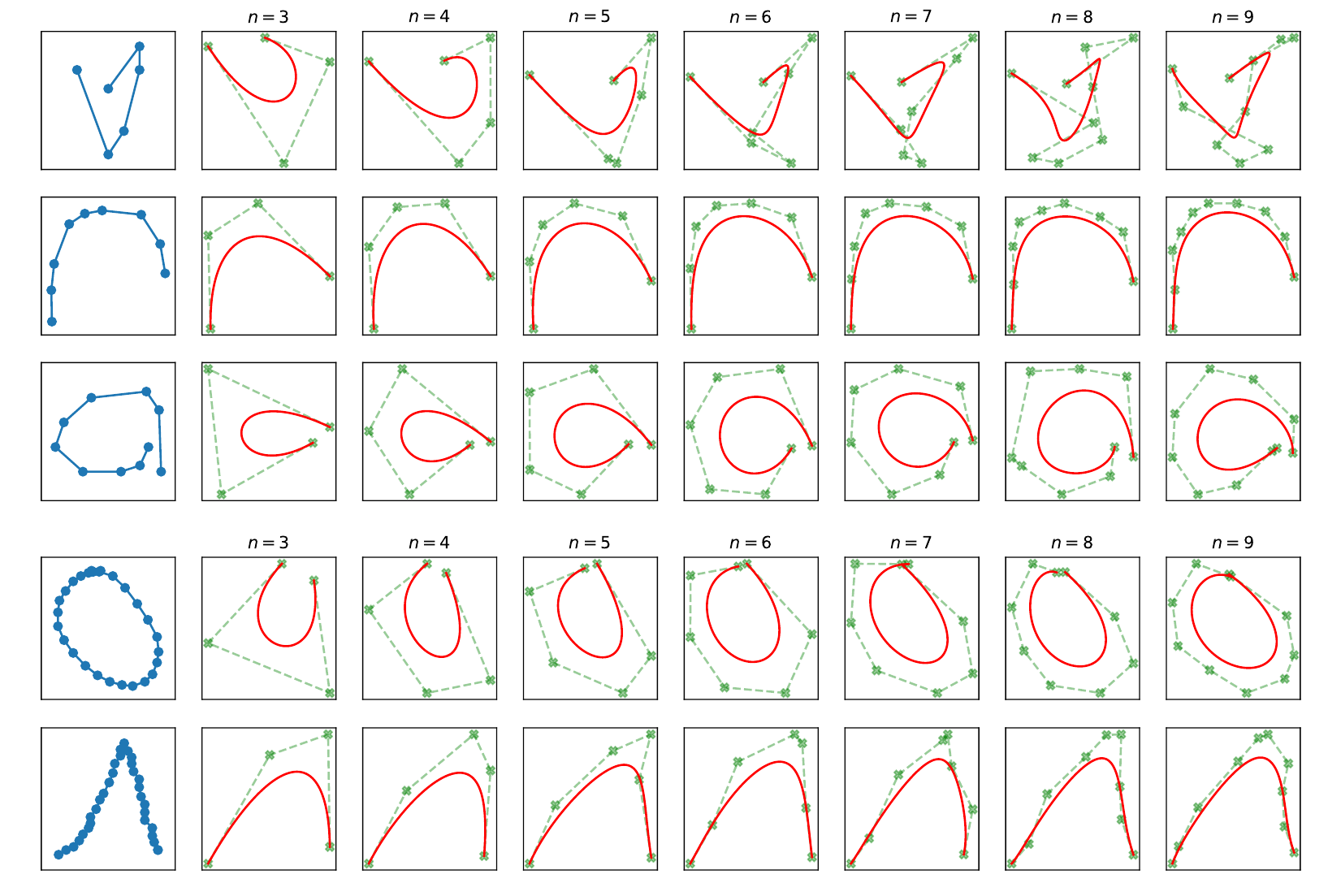}%
    \includegraphics[width=0.4\linewidth]{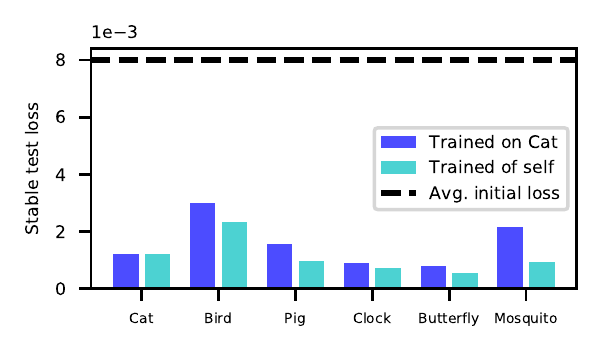}\\
    \includegraphics[width=0.98\linewidth]{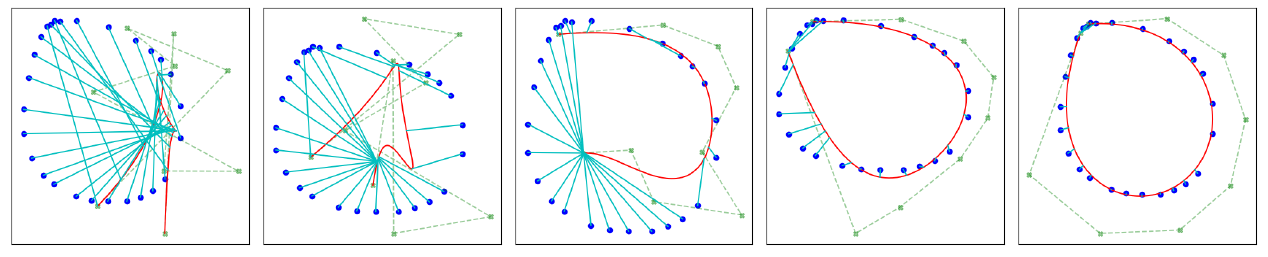}\\
    \includegraphics[width=0.98\linewidth]{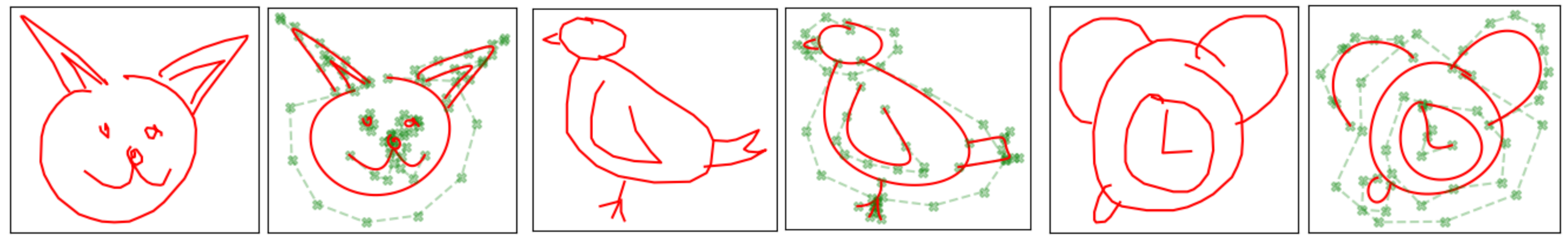}
    \caption{Evaluating our \bEnc{}. (Top left) Learned representations of multi-degree B\'ezier stroke embedding. Top and bottom rows contain moderate and high-sampling rate respectively. (Top right) Test loss for various categories when trained on same category vs ``Cat'', demonstrating transferability of the encoder. (Middle) Visualising training dynamics. Blue: Stroke to fit. Red and Green: B\'ezier curve and control points. Cyan: Estimated point correspondence. (Bottom) Examples of full sketches and their learned B\'ezier representation.}
    \label{fig:MultiDegreeFit}
    \vspace*{-2em}
\end{figure}


\keypoint{Implementation Details}
We created a dataset of all strokes from all sketches in a category of \emph{Quick, Draw!} in order to train the stroke embedding model described in Section~\ref{sec:StrokeEmbedding}. We adopted some tricks that made the training and representation more efficient in practice. We normalized all strokes to start from the origin (i.e., $\mathbf{X}_1 = [ 0, 0 ]^T $). 
Furthermore, we assumed that the first control point $\mathbf{P}_0$ of a B\'ezier representation is always aligned to the first absolute coordinate of the stroke (i.e., $\mathbf{X}_1 = \mathbf{P}_0$). Given these design choices, we can ignore the first control point (fixing it to origin) and only predict successive differences of control points (i.e., $\Delta \mathbf{P}_1 \triangleq \mathbf{P}_1 - \mathbf{P}_0$, $\Delta \mathbf{P}_2 \triangleq \mathbf{P}_2 - \mathbf{P}_1$ and so on) and then decode $\mathbf{P}_i$ as $\mathbf{P}_i = \sum_{i'=1}^i \Delta \mathbf{P}_{i'}$ while evaluating the loss in Eq.~\ref{eq:BezierLoss}. We chose the hidden state dimension to be $h = 256$ and $n_{min} = 3, n_{max} = 9$ for learning multi-degree B\'ezier representation. To exclude over complicated strokes, we apply some heuristics to split a stroke into two or more. Specifically, we split a stroke into multiple parts based on two criteria: 1. Every part is within a maximum length and 2. Every part has only one sharp bend (determined by computing its curvature at a given point).
We set the regularizer weight $\beta = 10^{-3}$.

\keypoint{Results}
We first qualitatively demonstrate the results of inferring B\'ezier representations of input strokes. 
Fig.~\ref{fig:MultiDegreeFit}(top left) shows fitting results for various curve orders (columns) -- showing variable amounts of detail being captured at different orders. It also shows  fitting examples at both low (above) and high (below) sampling rates -- confirming that our encoder can adapt to both.

We next qualitatively illustrate the training dynamics of our model via the fit estimated as training progresses. The results in Fig.~\ref{fig:MultiDegreeFit}(middle) show the estimated fit during training in terms of B\'ezier curve (red) and control points (green) for a stroke defined by (blue) points. Recall that our encoder also predicts the interpolation parameters $t$ that match each input point to a location on the curve. These correspondences are indicated in (cyan). Clearly both the fit and the estimated correspondences improve with training iterations.
Refer to Appendix C in the supplementary document for similar visualization of more samples.

Given that our training data is grouped into categories, we next verify that our encoder indeed learns a generic B\`ezier embedding, and is not overfitted to a specific category. Specifically, we compare the test loss for reconstructing data of each category when the encoder is trained on the same category as testing vs trained vs a disjoint category to testing. The results in Fig.~\ref{fig:MultiDegreeFit}(top right) shows that the embedding generalizes quite well to categories it is not trained on.

Finally, Fig.~\ref{fig:MultiDegreeFit}(bottom) shows examples of full sketches encoded by our encoder, and then decoded as B\'eziers. We can see that the encoded sketches reflect the input, but are smoother and cleaner.


\subsection{Sketch generation Experiments}

\keypoint{Setup}
{In control point mode, a fully trained multi-degree embedding model is used to restructure all sketches in our dataset as $\mathcal{S}_{cp}$. We set $\mathcal{L}_{tolerance} = 10^{-3}$ to select the best $n$.} We then train a SketchRNN-like model \cite{ha2017neural} using the restructured data. As data augmentation, we added 2D standard normal noise at all control points.
Sampling from the latent space and decoding it by the decoder will generate sequence of control points and stroke/sketch ending bits. Treating one entire stroke as a set of control points, we can then draw it on a canvas using Eq.~\ref{eq:BezierEq} with any required level of granularity.

In stroke mode, we encode each stroke with a fixed degree of $n=9$. Very similar to \emph{control point mode}, we use a Bi-LSTM to encode the whole sketch stroke-by-stroke and extract $N_z$ dimensional latent vector. By conditioning on the latent vector, the decoder produces B\'ezier representation $\mathcal{P}$ of one stroke at each time-step. Thus, the length of a sketch coincides with the number of strokes present in the sketch. At each step of the decoder, we sample one stroke from $p(\mathcal{P}_j\vert \mathbf{g}_j, \Theta)$ which is modeled as a GMM with $M=10$ mixture components. However, unlike the control point mode and its corresponding SketchRNN-like architecture, we do not use correlation parameter in the constituent Gaussians. This design choice makes the individual dimensions of the Gaussians independent, sampling from which is justified given property.~\ref{prop:SampleBezier}. Apart from $\mathcal{P}_j$, we predict one more quantity in practice: the start location $\mathbf{v}_j \triangleq ( v_x, v_y )^T_j$ of the stroke w.r.t the whole sketch. The need for $\mathbf{v}_j$ arises due to the practical consideration of relocating the start of each individual stroke at the origin while encoding them.

\begin{figure}[t]
    \centering
    \subfloat[]{{ \includegraphics[width=0.99\linewidth]{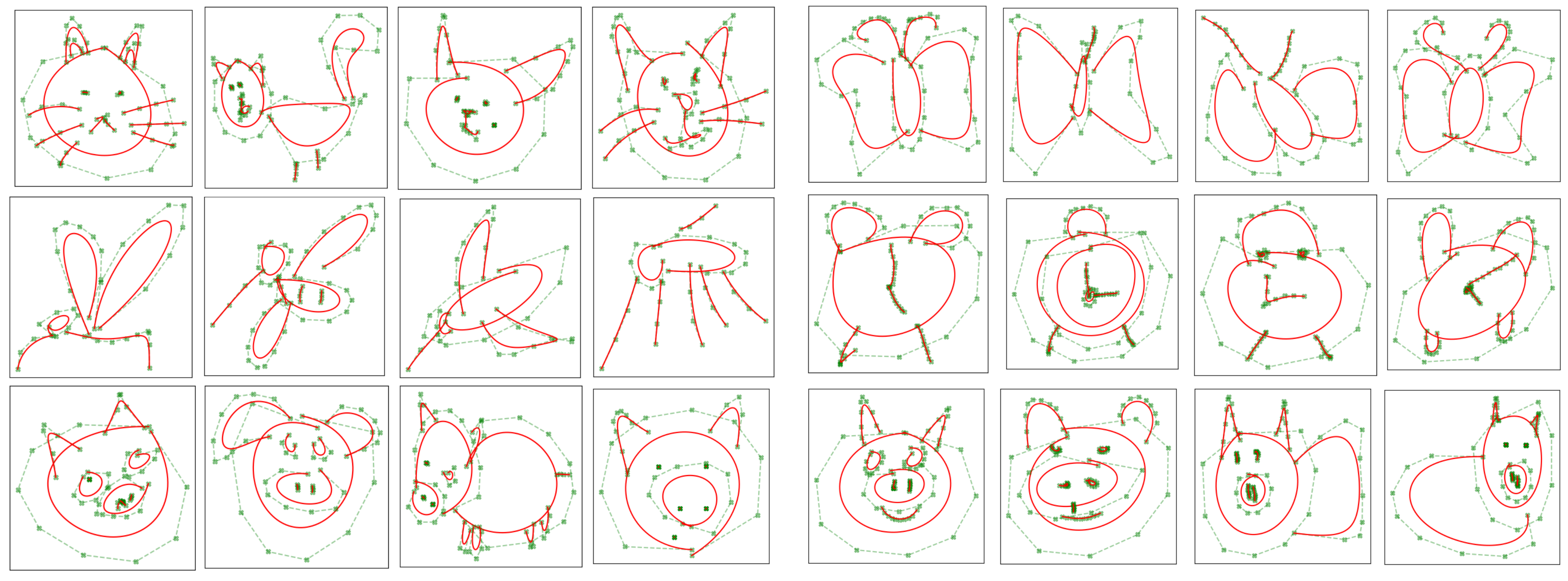} }}%
    \qquad
    \subfloat[]{{ \includegraphics[width=0.99\linewidth]{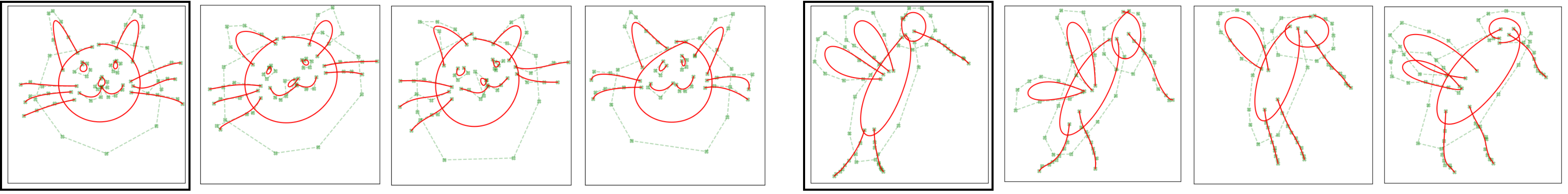} }}
    \caption{Qualitatively evaluating \bModel{}. (a) Samples drawn unconditionally in control point mode (left half) and stroke mode (right half). (b) Sketch samples generated by conditioning on the first sketch (double bordered) in each set.}
    \label{fig:OurGenSamples}
\end{figure}

\keypoint{Results}  Qualitative results of  generated unconditional sketch samples from both our model variants are shown in Fig.~\ref{fig:OurGenSamples}(a). We can see that, similarly to SketchRNN, \bModel{} generates diverse and plausible samples. However, uniquely our samples are high-resolution vector graphic sketches. Fig.~\ref{fig:OurGenSamples}(b) also shows examples of conditional samples where the right group of three images are samples conditioned on the left sketch encoding.

\begin{figure}[t]
    \centering
    \includegraphics[width=0.95\linewidth]{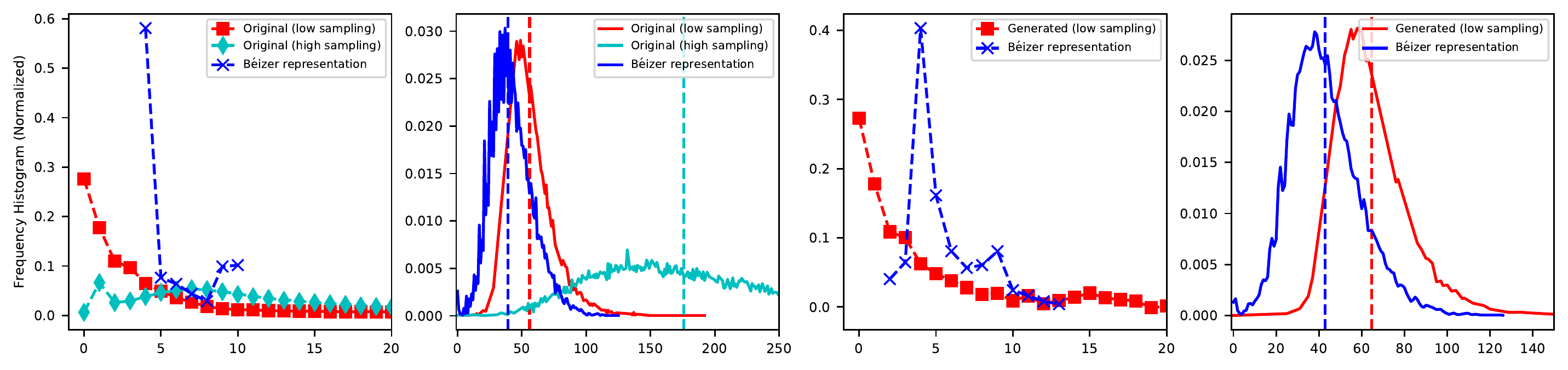}%
    \caption{Stroke/Sketch Length histogram for original data (left) and generated samples (right). B\'ezier encodings are shorter sequences than the raw data.}
    \label{fig:SketchLengthHisto}
\end{figure}

The use of B\'ezier curves as stroke representation reduces the average length of a given stroke's representation significantly and as a direct consequence, the description length for whole sketches as well. In Fig.~\ref{fig:SketchLengthHisto}, we compare the length histograms of original data and its B\'ezier representation both on stroke and sketch level, confirming that B\'eziers are systematically shorter (left). This is the same for strokes and sketches sampled by vanilla and SketchRNN and \bModel{} respectively (right).

This property of shorter representations for any given sketch means that our generator should have an advantage modeling longer sketches compared to vanilla SketchRNN since it only needs to model shorter sequences. To evaluate this, we use a modified Fr\'echet Inception Distance (FID) \cite{fidscore} score to compare the generated samples from both models. We first trained both our generator model and SketchRNN on the entire dataset (of each category). We then create a subset of sketches whose original length is $l\pm 20$ and use them to generate samples. All original and generated samples are rendered on a canvas and projected down to a concise feature vector using pre-trained Sketch-a-Net 2.0 \cite{sketchanet2} classifier. We compute the empirical mean  and covariance  of both real samples and generated samples as $(\mu_r, \Sigma_r$) and $(\mu_g, \Sigma_g)$ and then estimate modified FID as:
$$
\textsc{FID} = \norm{ \mu_r - \mu_g }^2 + \mathrm{Tr}( \Sigma_r + \Sigma_g - 2(\Sigma_r \Sigma_g)^{1/2})
$$

\begin{figure}
    \centering
    \includegraphics[scale=0.47]{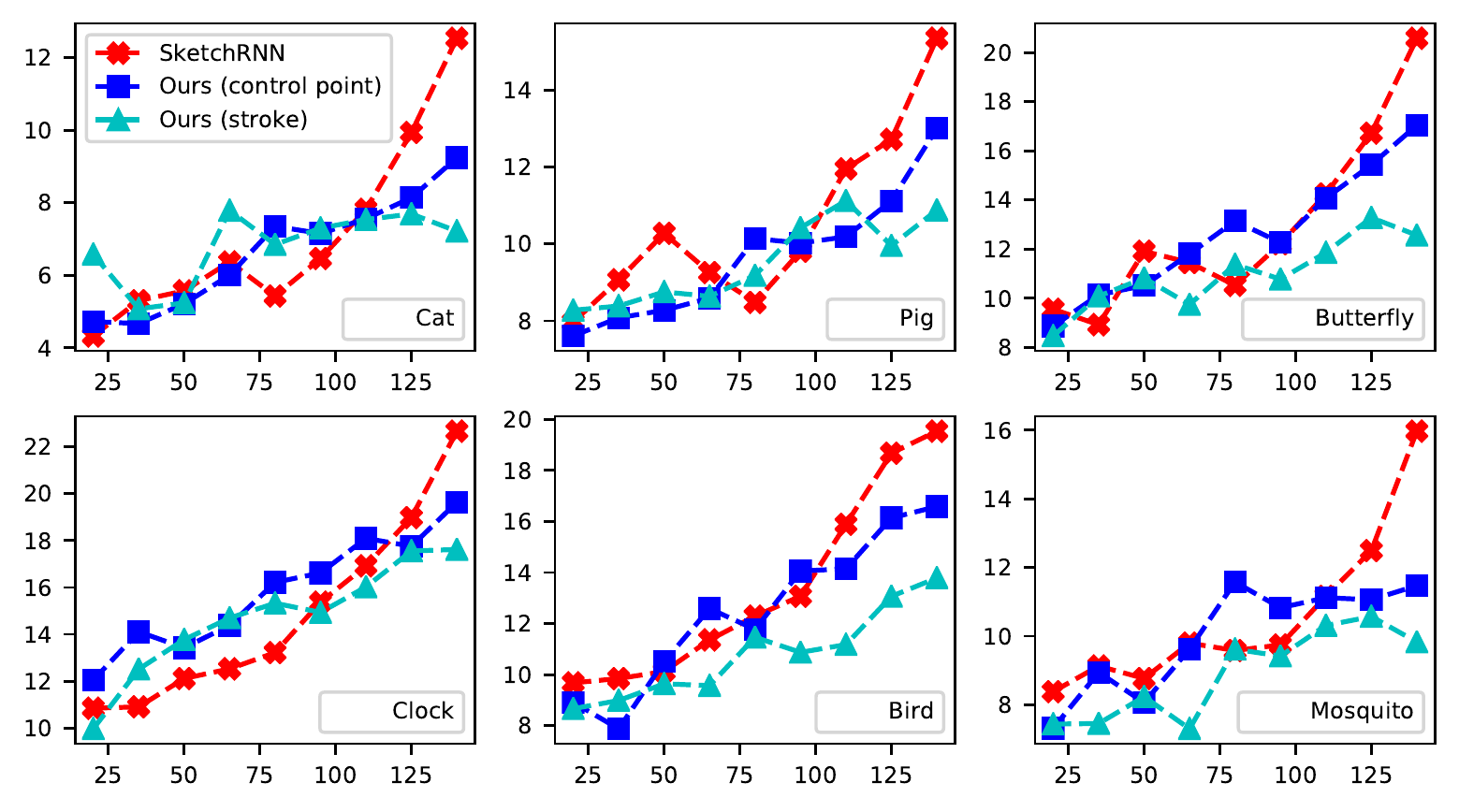}
    \includegraphics[scale=0.5]{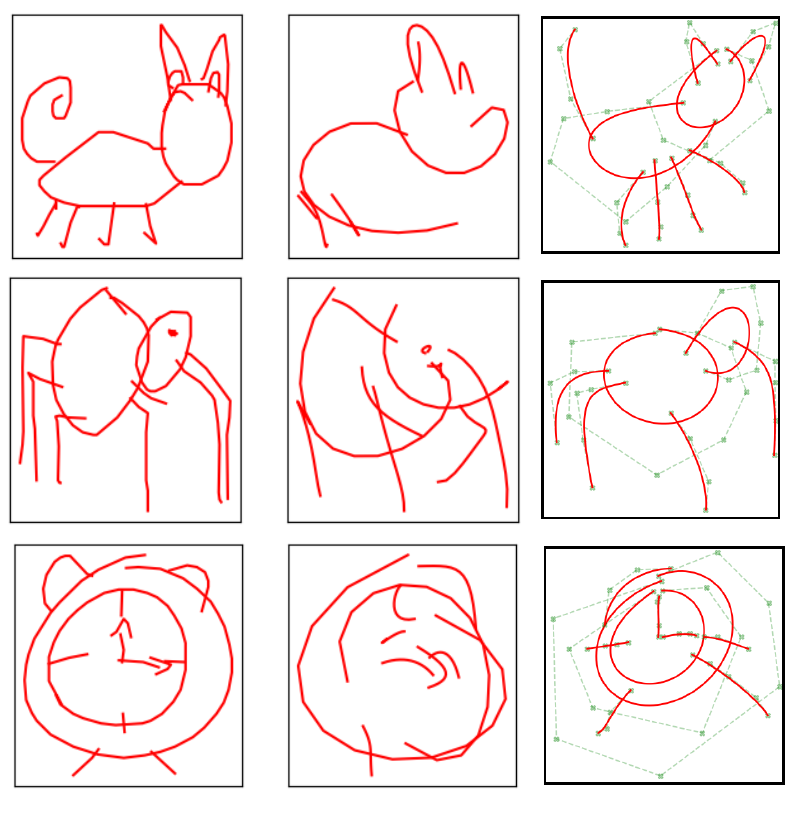}
    \vspace{-0.2cm}
    \caption{Left: FID score ($\downarrow$) vs length of sketch shows the effectiveness of our generative model on longer sketches. Right: Qualitative samples of long sketches. Three columns denote the original sketch, SketchRNN and our \bModel{}. }
    \label{fig:FIDComparison}
\end{figure}

The results in Fig.~\ref{fig:FIDComparison} plots the modified FID score with increasing length value $l$ for both SketchRNN and our model on each category of sketches. We can see that our model leads to improved (lower) FID score, especially for longer sketches. This is illustrated qualitatively in Fig.~\ref{fig:FIDComparison}, where we can see that for longer sketches, our framework produces much more reliable reconstruction than QuickDraw, which fails to make reasonable reconstruction in these cases.


\begin{figure}
    \centering
    \includegraphics[width=0.8\linewidth]{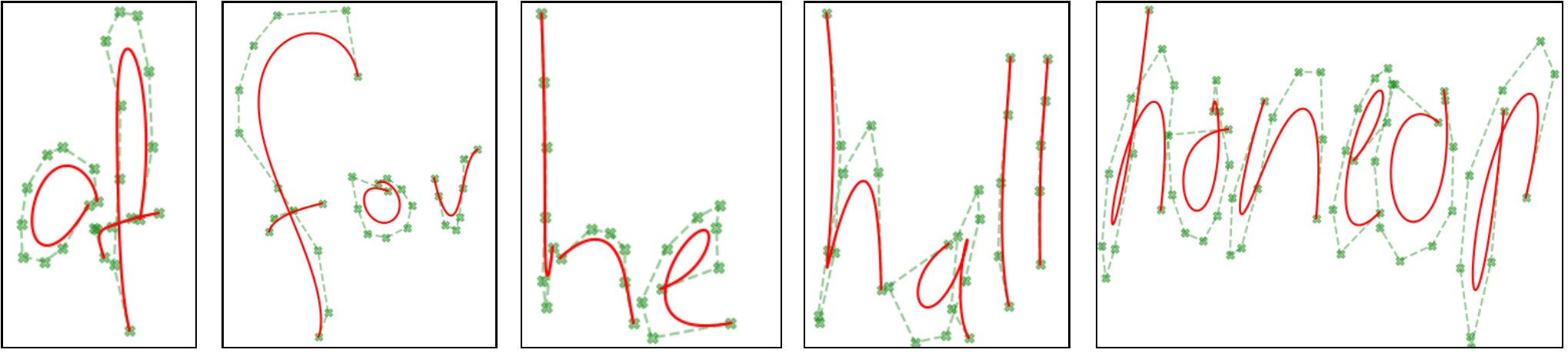}
    \vspace{-0.2cm}
    \caption{Unconditionally generating handwritten words from the IAM database.}
    \label{fig:iamsamples}
\end{figure}

\keypoint{Other applications}
Although crafted with sketches in mind, our framework can be adapted to other applications like handwriting generation (in line with the work of \cite{graves13}) with little to no modification. In fact, any 2D sequence data with two-level hierarchical representation (e.g., stroke and sketch) can be modeled using the same framework. Online handwritten characters are composed of relatively short strokes which we model with B\'ezier curves. We use the online handwritten sentences from the IAM handwriting database \cite{iamdataset}, embed the constituent strokes with our B\'ezier representation and train our generative model for words. Fig.~\ref{fig:iamsamples} shows qualitative samples from our resulting word generator.

\section{Conclusions}
In this paper we presented an inverse graphics approach to training an efficient model-based single-pass stroke-to-B\'ezier encoder via reconstruction through a B\'ezier decoder. Such approach surpasses the conventional fitting-based methods in terms of quality and efficiency. Furthermore, this enabled us to advance generative sketch models by generating sketches as sequences of parameterized curves rather than pixels, leading to \cut{generative models that provides} arbitrary-resolution scalable vector graphic samples. This new representation also enables better generation of longer sketches compared to existing state of the art. In future work we will investigate extending to more complex parameterized curves such as B-splines, and developing an encoder to predict curves from rasterized images directly.




\def\ECCVSubNumber{5496}  

\title{Supplementary material for \\ B\'ezierSketch: A generative model for scalable vector sketches} 

\titlerunning{B\'ezierSketch: A generative model for scalable vector sketches}

\author{
	Ayan Das\inst{1,2}\and
	Yongxin Yang\inst{1,2}\and
	Timothy Hospedales\inst{1,3}\and
	Tao Xiang\inst{1,2}\and
	Yi-Zhe Song\inst{1,2}
}

\authorrunning{A. Das et al.}

\institute{
	SketchX, CVSSP, University of Surrey, United Kingdom \\
	\email{\{a.das,yongxin.yang,t.xiang,y.song\}@surrey.ac.uk} \and
	iFlyTek-Surrey Joint Research Centre on Artificial Intelligence \and
	University of Edinburgh, United Kingdom \\
	\email{t.hospedales@ed.ac.uk}
}
\maketitle


\section{Appendix A}

\begin{property}
	\label{prop:SampleBezier}
	Given a $(\mathbf{T}, \mathcal{P})$ pair where $\mathbf{T} = \mathbf{d}(\mathcal{P})$ for an arbitrary set of $t$, and $\widehat{\mathcal{P}} \sim \mathcal{N}(\mathcal{P}, \Sigma)$, then the decoded $\widehat{\mathbf{T}} = \mathbf{d}(\widehat{\mathcal{P}})$ with the same set of $t$, is distributed as $\mathcal{N}(\mathbf{T}, \Sigma')$, where $\Sigma$ and $\Sigma'$ are diagonal covariance matrices.
\end{property}

\begin{proof}
	As $\Sigma$ is diagonal, we can separate each dimension of $\mathcal{N}(\mathcal{P}, \Sigma)$ into individual Gaussians and then group $x-y$ components of each control point with its own Gaussian with diagonal covariance $\Sigma_i \triangleq \begin{bmatrix} \sigma_{x_i}, 0 \\ 0, \sigma_{y_i} \end{bmatrix}$
	
	$$
	\mathcal{N}(\mathcal{P}, \Sigma) = \prod_{i=0}^n \mathcal{N}\left(\mathbf{P}_i, \Sigma_i \right)
	$$
	
	By drawing samples from the gaussians of individual control points, we get $\widehat{\mathcal{P}} \triangleq \left[ \widehat{\mathbf{P}}_i \right]_{i=0}^n$ where $\widehat{\mathbf{P}}_i \sim \mathcal{N}(\mathbf{P}_i, \Sigma_i)$. Decoding $\widehat{\mathcal{P}}$ by $\mathbf{d}(\cdot)$ gives
	
	\begin{equation}
	\widehat{\mathbf{T}} = \mathbf{d}(\widehat{\mathcal{P}}) = \sum_{i=0}^n \mathcal{B}_{i,n}(t) \cdot \widehat{\mathbf{P}}_i
	\end{equation}
	
	Given any value of $t=t$, the random variable $\widehat{\mathbf{T}}$ is a weighted sum of $n$ independent gaussian random variables with weights $\left[ \mathcal{B}_{i,n}(t) \right]_{i=0}^n$. Hence, $\widehat{\mathbf{T}}$ is distributed as
	
	\begin{equation}
	\widehat{\mathbf{T}} \sim \mathcal{N}\left( \sum_{i=0}^n \mathcal{B}_{i,n}(t) \cdot \mathbf{P}_i, \sum_{i=0}^n \mathcal{B}^2_{i,n}(t) \cdot \Sigma_i\right)
	\end{equation}
	
	Now we know that $\displaystyle{ \sum_{i=0}^n \mathcal{B}_{i,n}(t) \cdot \mathbf{P}_i \triangleq \mathbf{T} }$ and we denote $\displaystyle{ \sum_{i=0}^n \mathcal{B}^2_{i,n}(t) \cdot \Sigma_i \triangleq \Sigma' }$. So,
	
	$$
	\widehat{\mathbf{T}} \sim \mathcal{N}(\mathbf{T}, \Sigma')
	$$
	
\end{proof}

\section{Appendix B}

\subsubsection{Sketch-RNN} \cite{ha2017neural} is considered the state-of-the-art generative model for free-hand vector sketches. Sketch-RNN models the consecutive differences of 2D waypoints of a sketch along with three bits denoting ``touching", ``stroke-end" and ``sketch-end" state of the pen. In control point mode of B\'ezierSketch, we adopted the same architecture and data representation as Sketch-RNN but with control points instead of waypoints. Hence, a sketch $\mathcal{S}_{cp}$ is transformed to a list (of length $N$) of 5-tuples $\displaystyle{ s_i \triangleq (\Delta P_x, \Delta P_y, q_1, q_2, q_3)_i }$ where $\left[ \Delta P_x, \Delta P_y \right]^T \triangleq \Delta\mathbf{P}$ is the successive difference of control points and $(q_1, q_2, q_3) \triangleq q$ are the three flag bits described above. As a normalization step, all sketches have been assumed to start from the origin (i.e., $\left[ 0, 0 \right]^T$).

The core model of Sketch-RNN is a Sequence-to-Sequence Variational Autoencoder (Seq2Seq-VAE) \cite{srivastava2015unsupervised} with a standard sequence encoder and an autoregressive decoder. The whole sketch sequence is fed into a Bidirectional encoder LSTM with hidden state given as

\begin{equation}
\mathbf{h}_i \triangleq \left[ \overrightarrow{\mathbf{h}}_i; \overleftarrow{\mathbf{h}}_i \right] = \text{Bi-LSTM}(s_i, \mathbf{h}_{i-1})
\end{equation}

\noindent and the last state $\mathbf{h}_N$ is used as a compact representation of the sketch. $\mathbf{h}_N$ is then used to generate the parameters of a gaussian distribution following the VAE framework \cite{Kingma2013AutoEncodingVB}. A sample is then drawn from the distribution as

$$
\mathbf{z} \sim \mathcal{N}(\mu, \sigma)\text{, where } \left[ \mathbf{\mu}, \mathbf{\sigma} \right] = f(\mathbf{h}_N) \in \mathbb{R}^Z
$$

\noindent and decoded by an autoregressive decoder. An unidirectional LSTM is employed to initialize from $\mathbf{z}$ and produce a reconstruction of the sketch sequence similar to \cite{graves13}. At each time-step $j$ of the decoder, the hidden state is given as

$$
\mathbf{g}_j = \text{LSTM}(\left[ \mathbf{z}; s_j \right], \mathbf{g}_{j-1})\text{, with } \mathbf{g}_0 = \text{tanh}(\mathbf{z})
$$

The decoder, at every time-step, outputs the parameters of a GMM (with $M$ mixtures) on $\left[ \Delta P_x, \Delta P_y\right]^T$ and also a categorical distribution on three flag bits discussed above. Samples from these distributions are fed back as input $s_{j+1}$ at next time step

\vspace*{-1em}

\begin{equation}
\begin{split}
s'_j &= \bigl(\Delta \mathbf{P}'_j, q'_j \bigr)\text{, where } \\
\Delta \mathbf{P}'_j \sim \text{GMM} & (\Delta \mathbf{P}; \mathbf{g}_j)\text{ and }q'_j \sim \text{Cat}(q; \mathbf{g}_j)
\end{split}
\end{equation}

The network is trained with the following loss that comprises of log-likelihood of the GMM, categorical cross-entropy of the flag bits and a variational KL divergance loss

\vspace*{-1em}

\begin{equation}
\begin{split}
L = -\frac{1}{N_{max}} \left[ \sum_{j=1}^N \log \text{GMM}(\Delta \mathbf{P}'_j) + \sum_{j=1}^{N_{max}} q_j \log q'_j \right] \\
- \frac{1}{2Z} (1 + \sigma - \mu^2 - exp(\sigma))
\end{split}
\end{equation}

\section{Appendix C}

\begin{figure}
	\centering
	\includegraphics[width=0.85\linewidth]{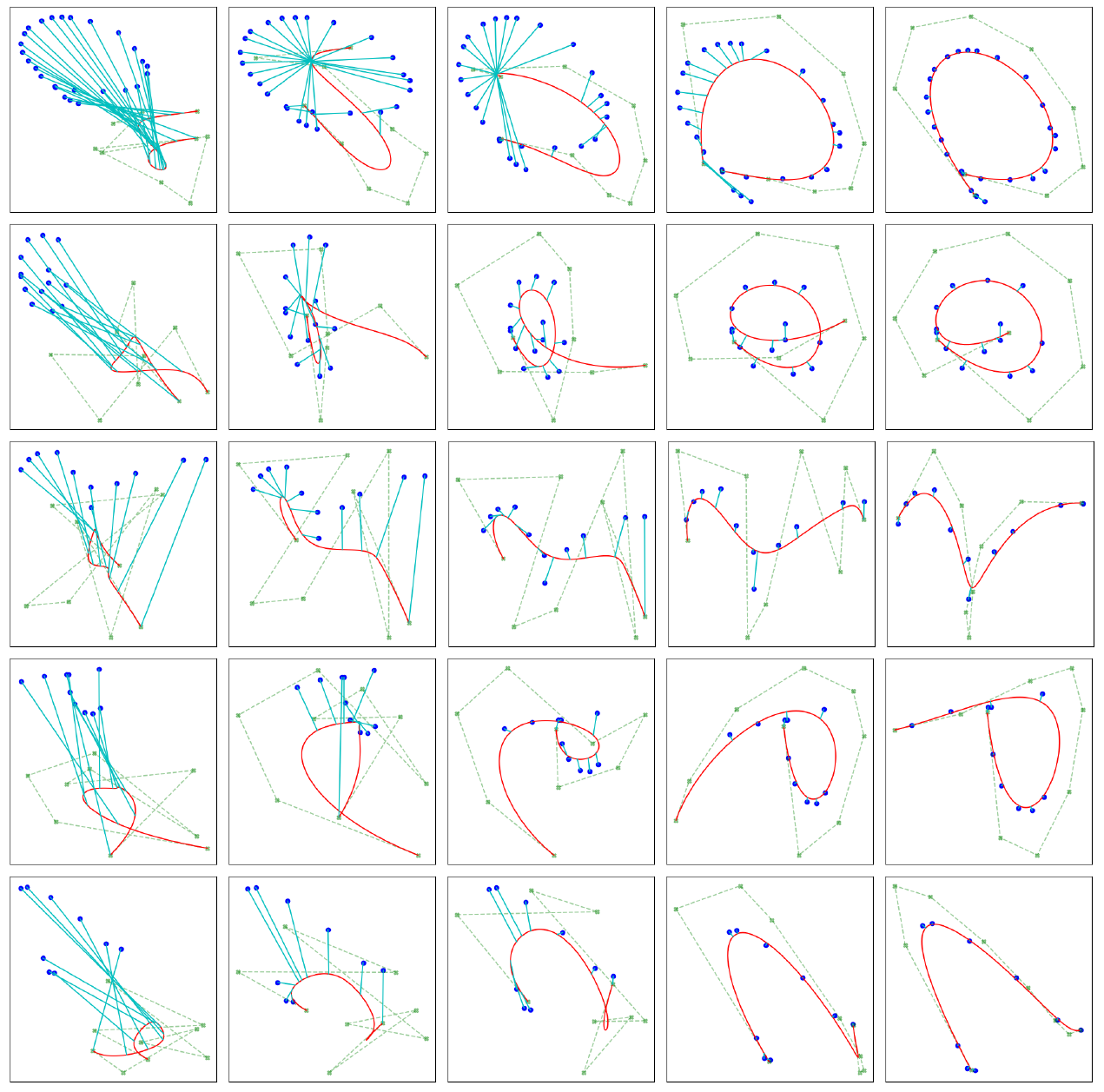}
	\caption{Visualization of intermediate stages of the fitting for B\'ezierEncoder network. Each row corresponds to one sample and columns denote increasing iterations of training.}
	\label{fig:optimviz}
\end{figure}

We provide visualizations (Refer to Fig.~\ref{fig:optimviz}) of the optimization dynamics over time. We also annotate a discrete point of the stroke and its corresponding point on the B\'ezier curve by joining them by a connector.

\bibliographystyle{splncs04} 
\bibliography{main}

\begin{thebibliography}{10}
\providecommand{\url}[1]{\texttt{#1}}
\providecommand{\urlprefix}{URL }
\providecommand{\doi}[1]{https://doi.org/#1}

\bibitem{bishop1994mixture}
Bishop, C.M.: Mixture density networks. Tech. rep., Aston University (1994)

\bibitem{bowman-etal-2016-generating}
Bowman, S.R., Vilnis, L., Vinyals, O., Dai, A., Jozefowicz, R., Bengio, S.:
  Generating sentences from a continuous space. In: CoNLL (2016)

\bibitem{deboor}
De~Boor, C., De~Boor, C., Math{\'e}maticien, E.U., De~Boor, C., De~Boor, C.: A
  practical guide to splines, vol.~27. Springer-Verlag New York (1978)

\bibitem{sketchx_fgsbir_2}
Dey, S., Riba, P., Dutta, A., Llados, J., Song, Y.Z.: Doodle to search:
  Practical zero-shot sketch-based image retrieval. In: CVPR (2019)

\bibitem{ganin2018imagePrograms}
Ganin, Y., Kulkarni, T., Babuschkin, I., Eslami, S.M.A., Vinyals, O.:
  Synthesizing programs for images using reinforced adversarial learning. In:
  ICML (2018)

\bibitem{GANgoodfellow14}
Goodfellow, I., Pouget-Abadie, J., Mirza, M., Xu, B., Warde-Farley, D., Ozair,
  S., Courville, A., Bengio, Y.: Generative adversarial nets. In: NIPS (2014)

\bibitem{graves13}
Graves, A.: Generating sequences with recurrent neural networks. CoRR
  \textbf{abs/1308.0850} (2013)

\bibitem{ha2017neural}
Ha, D., Eck, D.: A neural representation of sketch drawings. In: ICLR (2018)

\bibitem{fidscore}
Heusel, M., Ramsauer, H., Unterthiner, T., Nessler, B., Hochreiter, S.: Gans
  trained by a two time-scale update rule converge to a local nash equilibrium.
  In: NIPS (2017)

\bibitem{Hinton504}
Hinton, G.E., Salakhutdinov, R.R.: Reducing the dimensionality of data with
  neural networks. Science  \textbf{313}(5786),  504--507 (2006)

\bibitem{pix2pix_isola}
Isola, P., Zhu, J.Y., Zhou, T., Efros, A.A.: Image-to-image translation with
  conditional adversarial networks. In: CVPR (2017)

\bibitem{Kingma2013AutoEncodingVB}
Kingma, D.P., Welling, M.: Auto-encoding variational bayes. ICLR  (2014)

\bibitem{klare2011mugshotSketchMatch}
Klare, B., Li, Z., Jain, A.: Matching forensic sketches to mug shot photos.
  IEEE Transactions on Pattern Analysis and Machine Intelligence
  \textbf{33}(3),  639 --646 (march 2011)

\bibitem{kulkarni2015vig}
Kulkarni, T.D., Whitney, W., Kohli, P., Tenenbaum, J.B.: Deep convolutional
  inverse graphics network. In: NIPS (2015)

\bibitem{lake_bpl}
Lake, B.M., Salakhutdinov, R., Tenenbaum, J.B.: Human-level concept learning
  through probabilistic program induction. Science  \textbf{350}(6266),
  1332--1338 (2015)

\bibitem{laube2018bSplineNeural}
{Laube}, P., {Franz}, M.O., {Umlauf}, G.: Deep learning parametrization for
  b-spline curve approximation. In: 2018 International Conference on 3D Vision
  (3DV) (2018)

\bibitem{spliefit_persample1}
Liu, Y., Wang, W.: A revisit to least squares orthogonal distance fitting of
  parametric curves and surfaces. In: GMP (2008)

\bibitem{fontgen_iccv}
Lopes, R.G., Ha, D., Eck, D., Shlens, J.: A learned representation for scalable
  vector graphics. In: ICCV (2019)

\bibitem{iamdataset}
Marti, U.V., Bunke, H.: A full english sentence database for off-line
  handwriting recognition. In: ICDAR (1999)

\bibitem{advancecubicbezierfit}
Masood, A., Ejaz, S.: An efficient algorithm for robust curve fitting using
  cubic bezier curves. In: ICIC (2010)

\bibitem{sketchx_fgsbir_1}
Pang, K., Li, K., Yang, Y., Zhang, H., Hospedales, T.M., Xiang, T., Song, Y.Z.:
  Generalising fine-grained sketch-based image retrieval. In: CVPR (2019)

\bibitem{spliefit_persample2}
Plass, M., Stone, M.: Curve-fitting with piecewise parametric cubics. In:
  SIGGRAPH (1983)

\bibitem{rabiner_hmm}
{Rabiner}, L., {Juang}, B.: An introduction to hidden markov models. IEEE ASSP
  Magazine  \textbf{3}(1),  4--16 (1986)

\bibitem{dcgan}
Radford, A., Metz, L., Chintala, S.: Unsupervised representation learning with
  deep convolutional generative adversarial networks. In: ICLR (2016)

\bibitem{digit_splines_hinton}
Revow, M., Williams, C.K.I., Hinton, G.E.: Using generative models for
  handwritten digit recognition. IEEE Transactions on Pattern Analysis and
  Machine Intelligence  \textbf{18}(6),  592–606 (1996)

\bibitem{romaszko2017vig}
{Romaszko}, L., {Williams}, C.K.I., {Moreno}, P., {Kohli}, P.:
  Vision-as-inverse-graphics: Obtaining a rich 3d explanation of a scene from a
  single image. In: ICCVW (2017)

\bibitem{salomon2007curves}
Salomon, D.: Curves and surfaces for computer graphics. Springer Science \&
  Business Media (2007)

\bibitem{sangkloy2016sketchy}
Sangkloy, P., Burnell, N., Ham, C., Hays, J.: The sketchy database: Learning to
  retrieve badly drawn bunnies. In: SIGGRAPH (2016)

\bibitem{cubicbezierfit}
Shao, L., Zhou, H.: Curve fitting with bezier cubics. Graphical models and
  image processing  \textbf{58}(3),  223--232 (1996)

\bibitem{song2018learn2Sketch}
Song, J., Pang, K., Song, Y., Xiang, T., Hospedales, T.M.: Learning to sketch
  with shortcut cycle consistency. In: CVPR (2018)

\bibitem{srivastava2015unsupervised}
Srivastava, N., Mansimov, E., Salakhudinov, R.: Unsupervised learning of video
  representations using lstms. In: ICML (2015)

\bibitem{policygrad}
Sutton, R.S., McAllester, D.A., Singh, S.P., Mansour, Y.: Policy gradient
  methods for reinforcement learning with function approximation. In: NIPS
  (1999)

\bibitem{sketchanet2}
Yu, Q., Yang, Y., Liu, F., Song, Y.Z., Xiang, T., Hospedales, T.: Sketch-a-net:
  A deep neural network that beats humans. International Journal of Computer
  Vision  \textbf{122},  411–425 (2017)

\bibitem{sketchanet1}
Yu, Q., Yang, Y., Song, Y.Z., Xiang, T., Hospedales, T.: Sketch-a-net that
  beats humans. In: BMVC (2015)

\bibitem{bspline_lbfgs}
Zheng, W., Bo, P., Liu, Y., Wang, W.: Fast b-spline curve fitting by l-bfgs.
  Computer Aided Geometric Design  \textbf{29}(7),  448–462 (2012)

\end{thebibliography}

\end{document}